\documentclass[review]{elsarticle}

\usepackage{lineno,hyperref}
\modulolinenumbers[5]

\journal{arXiv}

\begin{document}

\begin{frontmatter}

\title{A Simple Robot Selection Criteria After Path Planning Using Wavefront Algorithm}
%\tnotetext[mytitlenote]{Fully documented templates are available in the elsarticle package on \href{http://www.ctan.org/tex-archive/macros/latex/contrib/elsarticle}{CTAN}.}
\author{Rajashekhar V S \fnref{Corresponding author}}
\address{Life Member, The Robotics Society, India and Association for Machines and Mechanisms, India}
\fntext[myfootnote]{Corresponding author}
\ead{vsrajashekhar@gmail.com}

\author{Dhaya C}
\address{Professor, Department of Computer Science Engineering, Adhiparasakthi Engineering College, Melmaruvathur, Tamil Nadu, India}
%\ead{dhaya@apec.edu.in}

\author{Dinakar Raj C K}
\address{Professor, Department of Mechanical Engineering, Adhiparasakthi Engineering College, Melmaruvathur, Tamil Nadu, India}
%\ead{ckd@apec.edu.in}

\author{Dharshan P, Mukesh Kumar S, Harish B, Ajith R and Kamaleshwaran K}
\address{Student, Department of Computer Science Engineering, Adhiparasakthi Engineering College, Melmaruvathur, Tamil Nadu, India}

%% Group authors per affiliation:
%\author{Elsevier\fnref{myfootnote}}
%\address{Radarweg 29, Amsterdam}
%\fntext[myfootnote]{Since 1880.}

%% or include affiliations in footnotes:
%\author[mymainaddress,mysecondaryaddress]{Elsevier Inc}
%\ead[url]{www.elsevier.com}
%
%\author[mysecondaryaddress]{Global Customer Service\corref{mycorrespondingauthor}}
%\cortext[mycorrespondingauthor]{Corresponding author}
%\ead{support@elsevier.com}
%
%\address[mymainaddress]{1600 John F Kennedy Boulevard, Philadelphia}
%\address[mysecondaryaddress]{360 Park Avenue South, New York}

\begin{abstract}
In this work we present a technique to select the best robot for accomplishing a task assuming that the map of the environment is known in advance. To do so, capabilities of the robots are listed and the environments where they can be used are mapped. There are five robots that included for doing the tasks. They are the robotic lizard, half-humanoid, robotic snake, biped and quadruped. Each of these robots are capable of performing certain activities and also they have their own limitations. The process of considering the robot performances and acting based on their limitations is the focus of this work. The wavefront algorithm is used to find the nature of terrain. Based on the terrain a suitable robot is selected from the list of five robots by the wavefront algorithm. Using this robot the mission is accomplished. 
\end{abstract}


\end{frontmatter}

%\linenumbers
\section{Introduction}
\label{sec_introduction}
The path planning is done to find paths that are collision free for the robot to move from the start point to goal point. When the path is planned one can decide what robot can be used to navigate in that path. At times there can exist multiple options that lead to difficulty in decision making. Therefore we need to have a proper decision making tree that will lead to the selection of best possible robot to accomplish the task.   

The generalized wavefront algorithm creates fine and safe paths with reduced path length when compared to A$^*$ and RRT algorithm \cite{wu2020mobile}. In the recent works, the wavefront algorithm is applied to explore 3D spaces in indoor environments \cite{tang2020autonomous}. In \cite{pal2011focused}, a focused wavefront algorithm is introduced where there is a cost function assigned to some waves. Therefore the wavefront algorithm is one of the best algorithms used in path planning for robots. 

In this work we use the wavefront algorithm to find the optimal path. While finding the path, the algorithm finds the nature of the obstacles that are present on the path. There are five robots namely robotic lizard, half-humanoid, robotic snake, biped and quadruped that are available to be used to navigate in the path generated. Of these five robots that are considered in this work, one or more robots will be suggested by the wavefront algorithm for navigation in the path that it generated. This will aid in accomplishing the task that needs to be done.

\section{Wavefront Algorithm}
\label{sec_wavefront}
\begin{figure}[h!]
\begin{center}
\includegraphics[scale=0.50]{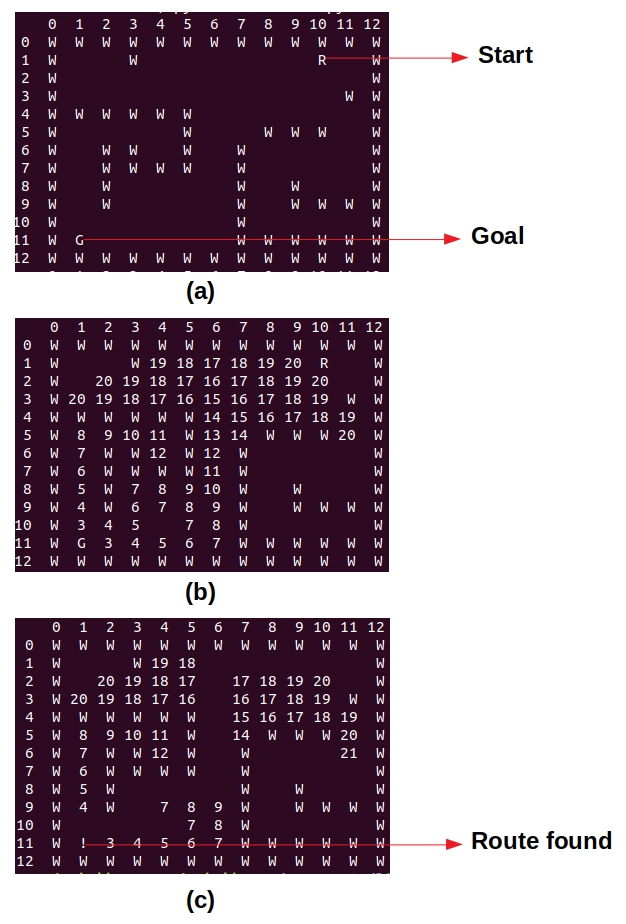}
\end{center}
\caption{The implementation of Wavefront algorithm (a) The goal and start position (b) The routes are generated (c) The optimal path is obtained}
\label{fig_algorithm}
\end{figure}

Wavefront algorithm is a breadth-first graph-search algorithm. It comes under deterministic graph search methods in robot path planning \cite{siegwart2011introduction}. The map is generated using one of the various methods available in the literature. The map consists of different terrains that are listed in Section \ref{subsec_terrain}. The robots used to travel in the map are mentioned in the Section \ref{subsec_robots_used}. The algorithm first starts searching from the goal position to the start position. Once the start point is reached, the shortest path to the goal position is generated. An example is shown in Figure \ref{fig_algorithm}.
 
\section{Robots and Nature of Terrain}
\label{sec_robots_used_nature_terrain}
Once the path is known, we need to find the nature of the terrain that the robot will have to travel.                  Based on this the robot will be selected to accomplish the task. The robots to be used and the nature of terrain are explained below. 
\subsection{Robots used}
\label{subsec_robots_used}
\begin{figure}[h!]
\begin{center}
\includegraphics[scale=0.350]{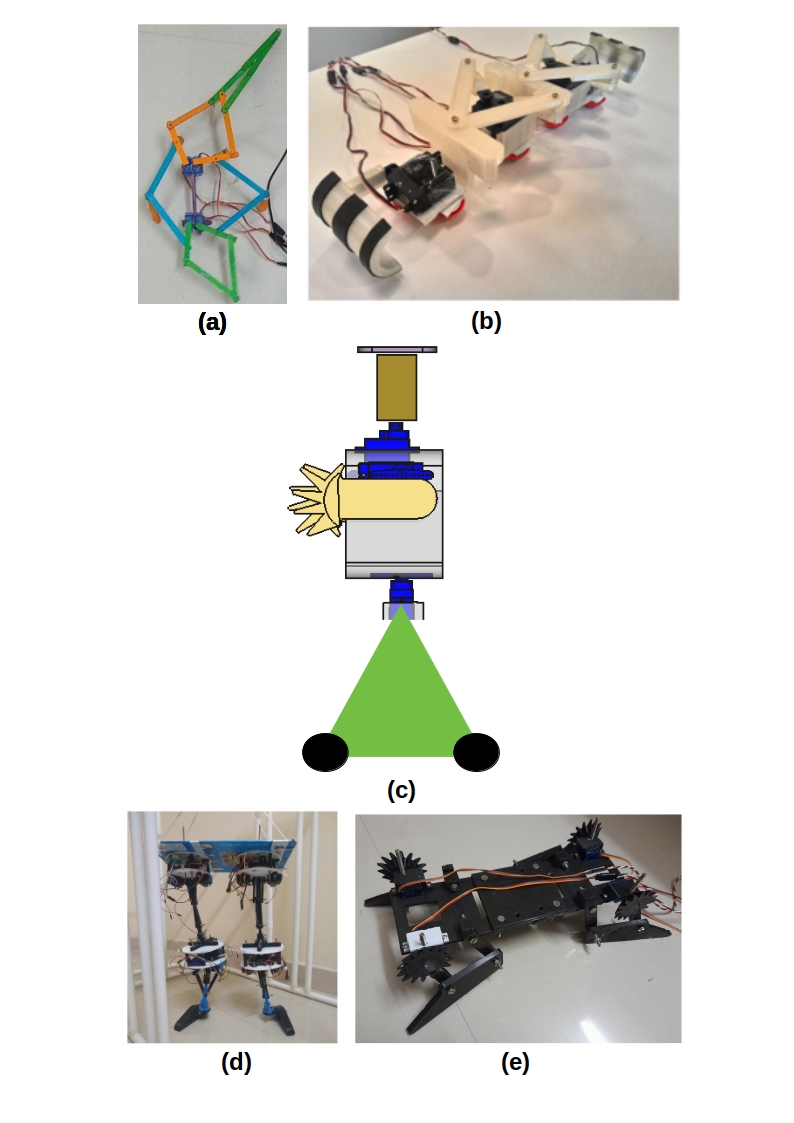}
\end{center}
\caption{The robots used for the mission (a) Robotic Lizard (b) Robotic Snake (c) Half-humanoid (d) Biped (e) Quadruped}
\label{fig_robots}
\end{figure}

There are five robots that are going to be used after the path has been decided. They are robotic lizard, half-humanoid, robotic snake, biped and quadruped. Most of the robots are built by the authors. The robotic lizard is based on four five-bar mechanisms \cite{rajashekhar2022design}. It is shown in Figure \ref{fig_robots} (a). The five-bar mechanism based snake robot \cite{rajashekhar2015serial, kumar2017robotic} is shown in Figure \ref{fig_robots} (b). The half-humanoid to be used is shown in Figure \ref{fig_robots} (c). The biped and quadruped are newly built by the authors and are shown in Figure \ref{fig_robots} (d) and (e). One of these robots or a combination of these robots are to be used for traversing the path that is planned. 

\subsection{Nature of Terrain}
\label{subsec_terrain}
The nature of the path determines the robot to be used. In this work we consider five types of terrains. They are walls, cluttered floor, stairs, slopes and flat surface. Based on the terrain or terrains detected by the wavefront algorithm, we can choose one robot, and distance from the start to goal position. In the future works we will also look into modifying the terrain or adding aids such as poles and ladders if possible.

\section{Robot Selection Criteria}
\label{sec_robot_selection_criteria}
\begin{figure}[h!]
\begin{center}
\includegraphics[scale=0.50]{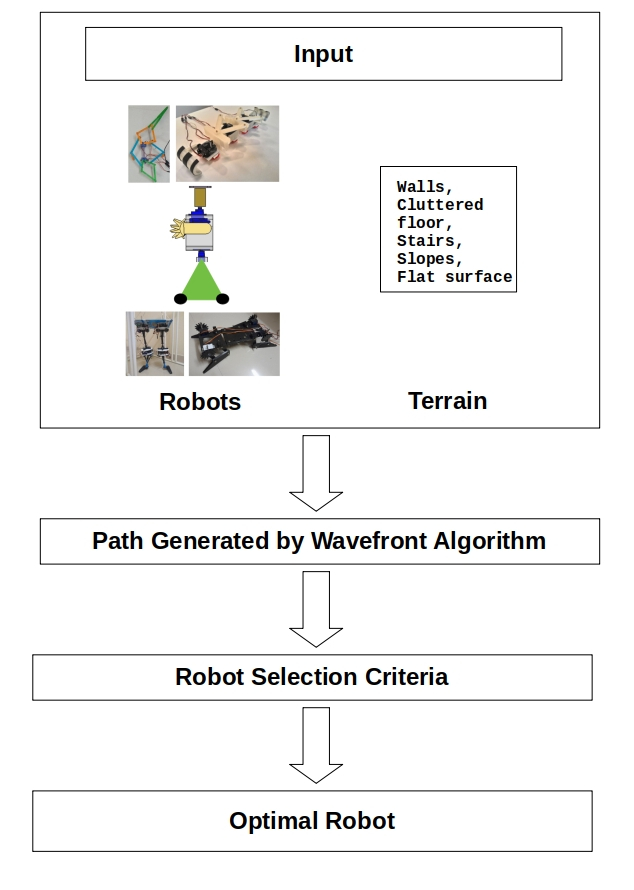}
\end{center}
\caption{The overall optimal robot selection process}
\label{fig_selection}
\end{figure}

The overall optimal robot selection process is shown in Figure \ref{fig_selection}. The robots that are available and the terrains where they can traverse are given as input to the path generated by the wavefront algorithm. The robot selection criteria is based on the terrain order of preference (\ref{equ_terrain}) and their equivalent robots in the order of preference (\ref{equ_robots}). The mapping between the robots and the terrain where they can be used are given in Table \ref{tab_selection}. 
 
\begin{equation}
\label{equ_terrain}
Wall > Stairs > Cluttered \hspace{1.5mm} floor > Slopes > Flat surfaces 
\end{equation}

\begin{equation}
\label{equ_robots}
Robotic \hspace{1.5mm} Lizard > Biped > Robotic \hspace{1.5mm} Snake > Quadruped > Half \hspace{1.5mm} Humanoid
\end{equation}

\begin{table}[h!]
\begin{center}
\caption{Robot selection based on the terrain}
\vspace{3mm}
\label{tab_selection}
\begin{tabular}{|c|c|}
\hline
\textbf{Terrain} & \textbf{Robot} \\ \hline
Wall             & Robotic Lizard \\ \hline
Stairs           & Biped          \\ \hline
Cluttered floor  & Robotic Snake  \\ \hline
Slope            & Quadruped      \\ \hline
Flat surface     & Half Humanoid  \\ \hline
\end{tabular}
\end{center}
\end{table}

\section{Implementation}
\label{sec_implementation}
\begin{figure}[h!]
\begin{center}
\includegraphics[scale=0.35]{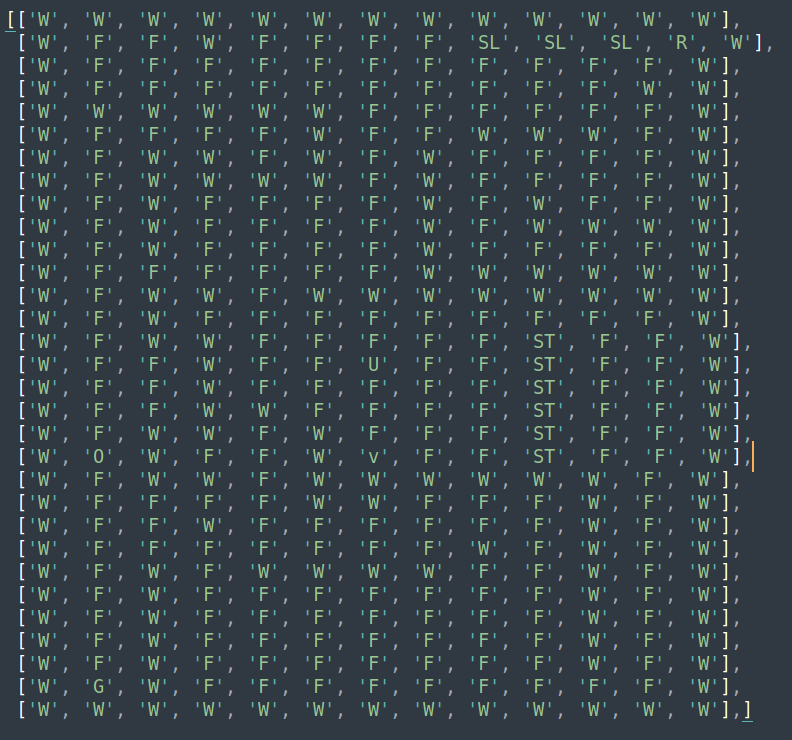}
\end{center}
\caption{The map that is fed to the algorithm to select the best robot}
\label{fig_given_map}
\end{figure}

If there is wall on the path, no matter what else terrain is present, the robotic lizard is used. If  stairs, cluttered floor, slope and flat surface is present, then biped is used to do the task. If the floor is cluttered, the robotic snake is used to traverse on it. In case there are slopes, the quadrupeds can be used to accomplish the task. Half humanoid can be used when the floor is perfectly flat. A sample map containing all the terrain is shown in Figure \ref{fig_given_map}.

\section{Conclusions}
In this work we presented a simple robot selection criteria after path planning. In future, we would add more terrains and robots. The power consumed by the robots would also be considered. Adding aids like ladders, poles and ropes will be done to  ease the movement of robots. Thus we hope that this simple method can be used during mission planning in order to choose the optimal robot needed to execute the task.
\section*{References}

\bibliography{mybibfile}

\begin{thebibliography}{1}
\expandafter\ifx\csname url\endcsname\relax
  \def\url#1{\texttt{#1}}\fi
\expandafter\ifx\csname urlprefix\endcsname\relax\def\urlprefix{URL }\fi
\expandafter\ifx\csname href\endcsname\relax
  \def\href#1#2{#2} \def\path#1{#1}\fi

\bibitem{wu2020mobile}
S.~Wu, Y.~Du, Y.~Zhang, Mobile robot path planning based on a generalized
  wavefront algorithm, Mathematical Problems in Engineering 2020 (2020) 1--12.

\bibitem{tang2020autonomous}
C.~Tang, Y.~Liang, S.~Yu, R.~Sun, J.~Zheng, Autonomous 3d exploration of indoor
  environment based on wavefront algorithm, in: 2020 IEEE International
  Conference on Networking, Sensing and Control (ICNSC), IEEE, 2020, pp. 1--6.

\bibitem{pal2011focused}
A.~Pal, R.~Tiwari, A.~Shukla, A focused wave front algorithm for mobile robot
  path planning, in: Hybrid Artificial Intelligent Systems: 6th International
  Conference, HAIS 2011, Wroclaw, Poland, May 23-25, 2011, Proceedings, Part I
  6, Springer, 2011, pp. 190--197.

\bibitem{siegwart2011introduction}
R.~Siegwart, I.~R. Nourbakhsh, D.~Scaramuzza, Introduction to autonomous mobile
  robots, MIT press, 2011.

\bibitem{rajashekhar2022design}
V.~Rajashekhar, C.~Dinakar~Raj, S.~Vishwesh, E.~Selva~Perumal, M.~Nirmal~Kumar,
  Design and analysis of a robotic lizard using five-bar mechanisms, in:
  Machines, Mechanism and Robotics: Proceedings of iNaCoMM 2019, Springer,
  2022, pp. 33--41.

\bibitem{rajashekhar2015serial}
V.~Rajashekhar, S.~Kumar, A serial five-bar mechanism based robotic snake
  exhibiting three kinds of gait, in: 2015 IEEE International Conference on
  Robotics and Biomimetics (ROBIO), IEEE, 2015, pp. 1938--1943.

\bibitem{kumar2017robotic}
S.~Kumar, R.~V. Saminathan, Robotic snake, uS Patent 9,796,081 (Oct.~24 2017).

\end{thebibliography}

\end{document}